\pdfoutput=1

\documentclass[11pt]{article}

\usepackage{float}
\restylefloat{table}
\usepackage{multirow}
\usepackage{makecell}
\usepackage{array,multirow,graphicx}
\usepackage{acl}

\usepackage{times}
\usepackage{latexsym}
\usepackage[T1]{fontenc}

\usepackage[utf8]{inputenc}

\usepackage{microtype}

\usepackage{arabtex}
\usepackage{utf8}
\setcode{utf8}
\usepackage{tipa}
\usepackage{inconsolata}
\usepackage{amsmath}

\usepackage{xspace}
\newcommand{\Ar}[1]{{\scriptsize \<#1>\xspace}}
\newcommand{\TrAr}[1]{\arabtrue\transfalse {\scriptsize \Ar{#1}} \arabfalse\transtrue \RL{#1}\arabtrue\transfalse}

\usepackage{abstract}
\title{Context-Gloss Augmentation for Improving Arabic Target Sense Verification}

\author{Sanad Malaysha, Mustafa Jarrar, Mohammed Khalilia \\
   Birzeit University, Palestine \\
    \texttt{\{smalaysha, mjarrar, mkhalilia\}@birzeit.edu} \\
 }

\begin{document}
\maketitle
\begin{abstract}
Arabic language lacks semantic datasets and sense inventories. The most common semantically-labeled dataset for Arabic is the \textit{ArabGlossBERT}, a relatively small dataset that consists of 167K context-gloss pairs (about 60K positive and 107K negative pairs), collected from Arabic dictionaries. This paper presents an enrichment to the ArabGlossBERT dataset, by augmenting it using (Arabic-English-Arabic) machine back-translation. Augmentation increased the dataset size to 352K pairs (149K positive and 203K negative pairs). We measure the impact of augmentation using different data configurations to fine-tune BERT on target sense verification (TSV) task. Overall, the accuracy ranges between 78\% to 84\% for different data configurations. Although our approach performed at par with the baseline, we did observe some improvements for some POS tags in some experiments. Furthermore, our fine-tuned models are trained on a larger dataset covering larger vocabulary and contexts. We provide an in-depth analysis of the accuracy for each part-of-speech (POS).
\end{abstract}

\section{Introduction}
\label{sec:introduction}
There are three tasks in the literature that are related to semantic understanding of natural language: \textit{(i)} Word Sense Disambiguation (WSD), \textit{(ii)} Target Sense Verification (TSV), and \textit{(iii)} Word-in-Context (WiC). WSD is the most common task, which aims to disambiguate word's semantics. Given a context (i.e., sentence), a target word in the context, and a set of candidate senses (i.e., glosses, meaning definitions \cite{J06}) for the target word, the goal of the WSD task is to determine \textit{which of these senses} is the intended meaning for the target word (\citealp{al2022arabglossbert}). For example, the word (\TrAr{جداول}) has two senses in Arabic: \textit{tables} (\Ar{شكل يحتوي على مجموعة قضايا أو معلومات }) and \textit{creek} (\Ar{ مجرى صغير متفرّع من نهر}). Thus, in the context (\Ar{تمشى بين الجداول والازهار}), WSD aims to determine which of the two senses is the intended meaning of (\TrAr{الجداول}). The TSV task is newly proposed in the literature (\citealp{breit2020wic}). It aims to classify a sentence pair with positive or negative. Given a context, target and gloss, TSV aims to decide whether this gloss is the intended meaning of the target. In other words, TSV does not determine which sense is the intended meaning, but rather, decides whether the context-gloss pair match (Positive) or not (Negative). For example, the sentence pair (\Ar{مجرى صغير متفرّع من نهر} - \Ar{تمشى بين الجداول والازهار}) is Positive, while the pair (\Ar{ شكل يحتوي على مجموعة قضايا أو معلومات} - \Ar{تمشى بين الجداول والازهار }) is Negative. WiC aims to determine whether a target word in two contexts is used in the same sense or not (\citealp{moreno2021ctlr}). Although the three tasks are closely related, they are not the same, and the choice of which task to use depends on the application scenario (e.g., machine translation, information retrieval, semantic tagging, or others). Some researchers try to address these tasks using different approaches. For example, \citet{hauer2021wic} proposed to solve the WSD by re-formulating it as a TSV task, a WiC task and a combination of TSV and WiC tasks.

Such semantic understanding tasks have been challenging for many years, but recently gained attention due to the advances in contextualized word embedding models \cite{al2022arabglossbert,HJ21}. Language models, specially BERT \cite{kenton2019bert}, have made significant advancements in down-streaming NLP tasks. BERT is a transformer-based model pre-trained on huge corpora \cite{devlin-etal-2019-bert}. It can be fine-tuned on domain/task-specific data (e.g., POS tagging, WSD, TSV, and WiC) to update its contextualized embeddings. The TSV task has been addressed by fine-tuning BERT on context-gloss pairs as a sentence pair binary classification problem (\citealp{huang2019glossbert}; \citealp{yap2020adapting}; \citealp{patel2021building}; \citealp{ranjbar2021lotus}; \citealp{lin2021context}; \citealp{el2021arabic}; \citealp{al2022arabglossbert}). However, the TSV task, similar to most NLP tasks, suffers from the knowledge-gain bottleneck, i.e., the lack of available quality datasets to train machine learning models.

Arabic is a low-resourced language \cite{DH21, JKG22} and the only available context-gloss pairs dataset is ArabGlossBERT \cite{al2022arabglossbert}. It consists of 167K context-gloss pairs, a relatively small dataset for fine-tuning BERT on a TSV task. The positive pairs (60K) were collected from multiple Arabic dictionaries \cite{JA19,J18,JAM19,J20} as well as from the Arabic Ontology \cite{J21,J11}. The pairs were cross-related to generate 106.8K negative pairs and used to fine-tune a BERT model, which achieved 84\% accuracy.

This paper aims to enrich the ArabGlossBERT dataset by augmenting it using the back-translation technique, similar to the work done for English by \citet{lin2021context}. With data augmentation, we generate new Arabic paraphrased contexts and glosses by translating the original data into English and back to Arabic, using Google Translate API. The new augmented dataset consists of 352K context-gloss pairs. To answer the question of whether the back-translation enrichment improves the TSV accuracy, we conduct 13 experiments that compare the accuracy obtained using the original dataset with the accuracy obtained using different combinations of the augmented datasets. We, also, provide an in-depth analysis of the TSV accuracy for each part-of-speech, which was not provided in \cite{al2022arabglossbert}. The main contributions of this work are:
\begin{itemize}
  \item Augmented ArabGlossBERT using back-translation (352K pairs).
  \item Thirteen experiments with different dataset configurations - to measure whether the back-translation enrichment can improve TSV  performance.
  \item In-depth analysis of the TSV accuracy for each part-of-speech.
\end{itemize}

The rest of the paper is organized as follows: Section \ref{sec:related_work} reviews the related literature, Section \ref{sec:data_augmentation} describes the data augmentation, Section \ref{sec:experiments} presents the experiments, Section \ref{sec:discussionconclusion} presents the results and we conclude in Section \ref{sec:future_work} with limitations and future work.

\section{Related Work}
\label{sec:related_work}
TSV has proven to be an effective solution for the WSD in many state-of-the-art efforts. Although some researchers did not use the term TSV, this notion was also referred to as GlossBERT or \textit{Context-Gloss Binary Classification} \cite{al2022arabglossbert,el2021arabic}. A TSV training dataset is typically a set of context-gloss pairs, each labeled with Positive or Negative. A pre-trained language model can then be fine-tuned for sentence pair binary classification. This idea was first proposed for English as GlossBERT (\citealp{huang2019glossbert}), where the training pairs were generated from a known SemCor dataset (\citealp{miller1993semantic}) with the target word, in context, marked up by a BERT-specific signal to emphasize it in the learning phase.

A similar effort in \cite{lin2021context} followed the GlossBERT technique. Their addition is the use of back-translation for improving the English WSD. They used back-translation from English to German and back to English in order to bridge the knowledge-gain gap and provide more context-gloss pairs. They also used a mark-up signal to surround the target word with double quotations. Only 2\% improvement was achieved using back-translation. This paper aims to evaluate this idea for Arabic. Another idea was proposed in \cite{yap2020adapting}, in which a learning signal (special token [TGT]) was used, and BERT was fine-tuned on sequence-pair ranking, the model selects the most related gloss given a context sentence and a list of candidate glosses. \citet{botha2020entity} used different mark-up signals in the form of open and close tags to emphasize the target word [E]target[/E] within the context sentence. 

For Arabic, the TSV task was addressed in \cite{al2022arabglossbert}, which presents the ArabGlossBERT, a dataset of 167K context-gloss pairs labeled with Positive or Negative. First, 60K positive pairs were extracted from different Arabic lexicons, then 106K negative pairs were generated automatically by cross-relating the positive pairs. The target word was marked-up with different learning signals. Different Arabic pre-trained models were fine-tuned, and the best model using AraBERT-V2 \cite{antoun2020arabert} achieved 84\% accuracy. Similar work for Arabic was proposed in \cite{el2021arabic} using a smaller dataset (15K positive and 15K negative) in which they used AraBERT-V2 and reported 89\% F1-score but this performance was criticized in \cite{al2022arabglossbert}.

\begin{table*}[ht]
\centering
\begin{tabular}{lrcl}
\hline
 &\textbf{Gloss} & - & \textbf{Context} \\ \hline
\makecell[l]{Original\\(Arabic)} & \Ar{فكرة أو مسألة تقدم للبحث} &- &  \Ar{جلس المسؤولون يناقشون \textbf{أطروحات} المشروع} \\ \hline

\makecell[l]{Translated\\(Arabic to English)} & \makecell[r]{An idea or question is\\ progressing to research} & - &\makecell[l]{Officials sat discussing\\ project proposals}\\ \hline

\makecell[l]{Back-Translated\\(English to Arabic)} & \Ar{فكرة أو سؤال مقدم للبحث} & - & \Ar{جلس المسؤولون لمناقشة مقترحات المشاريع} 
\\
\hline

\end{tabular}
\caption{\label{back-translation}
Example of context-gloss back-translation (Arabic-English-Arabic).}
\end{table*}

\section{Data Augmentation}
\label{sec:data_augmentation}

NLP tasks, including TSV, typically suffer from knowledge acquisition. The importance of knowledge acquisition is increasing especially because most NLP tasks are currently tackled using pre-trained neural models such as BERT, which generally requires large data to fine-tune. If the training data is not sufficient, the model will encounter the problem of unseen vocabulary and contexts, which harms model accuracy (\citealp{bevilacqua2021recent}). The linguistic resources that can be utilized for semantic understanding tasks are limited in Arabic language. Our assumption, for the TSV task, is that the more context-gloss pairs can be used during the training phase, the more vocabulary and more contexts will be covered, thus the better TSV accuracy. This is why researchers started to try new techniques for data augmentation in order to enrich the available dataset with more knowledge \cite{lin2021context,ranjbar2021lotus}.

For Arabic, and in order to enrich existing Arabic datasets, we propose to use the Arabic-English-Arabic back-translation, as illustrated in Table~\ref{back-translation}. It shows how the back-translation of glosses and contexts generates new paraphrased sentences with the same meaning. For back-translation we used Google Translate API, which was found to produce good quality and generally acceptable translations (\citealp{al2020diachronic}).  We did not remove diacritics since Arabic is diacritic-sensitive \cite{JZAA18}. Nevertheless, there are sentences that appeared with wrong or bad-quality translations, which we will discuss later. The translation was done in two phases. The glosses and contexts were translated into English, then back to Arabic. We, then, combined both the original dataset and the back-translated set. 

We only back-translated the ArabGlossBERT training dataset (152,035 pairs). The testing dataset (15,172 pairs) was not back-translated, because it is used as an evaluation benchmark to compare the performance improvement between the original and augmented datasets.

\begin{table*}[ht]
\centering
\begin{tabular}{lccr}
\hline
  & \textbf{\makecell{Original \\ ArabGlossBERT}} & \textbf{\makecell{Back-Translation \\Pairs}} & \textbf{\makecell{Augmented\\ArabGlossBERT}}\\
\hline
Unique un-diacritized lemmas & 26,169 & -- & 26,169\\
Unique glosses & 32,839 & 32,839 & 65,678\\
Unique contexts & 60,272 & 60,272 & 120,544\\
\textbf{Training pairs} & 152,035 & 152,035 + 32,839 & 336,909\\
\hspace{5pt}Positive pairs & 55,585 & 55,585 + 32,839 & 144,009\\
\hspace{5pt}Negative pairs & 96,450 & 96,450 & 192,900\\
\textbf{Testing pairs} & 15,172 & -- & 15,172\\
\hspace{5pt}Positive pairs & 4,738 & -- & 4,738\\
\hspace{5pt}Negative pairs & 10,434 & -- & 10,434\\
\textbf{Total: Train+Test} & 167,207 & 152,035 + 32,839 & 352,081\\
\hline
\end{tabular}
\caption{\label{original-dataset} Statistics of the original and augmented datasets.}
\end{table*}

Table~\ref{original-dataset} provides statistics about the original ArabGlossBERT dataset, the newly added back-translations, and the whole dataset after augmentation. The augmentation shows that the  size of the original dataset was doubled as it contains the original context-gloss pairs and the translated context-gloss pairs (152,032).

\begin{table*}
\centering
\begin{tabular}{|llrrr|}
\hline
\textbf{Dataset} & \textbf{Description} & \textbf{\makecell[l]{\small Positive \\ Pairs}} & \textbf{\makecell[l]{\small Negative \\ Pairs}} & \textbf{Total}\\
\hline

$D_1$ & The original ArabGlossBERT dataset & 55,585 & 96,450 & 152,035\\
\hline

$D_2$ & $D_1$ with target signal & 55,585 & 96,450 & 152,035\\
\hline

$D_3$ & $D_1$ with context replaced by back-translated context & 55,585 & 96,450 & 152,035\\
\hline

$D_4$ & $D_1$ + Positive pairs of $D_3$ & 111,170 & 96,450 & 207,620\\
\hline

$D_5$ & $D_1$ + $D_3$ & 111,170 & 192,900 & 304,070\\
\hline

$D_6$ & $D_1$ + Positive pairs {\small (original gloss - back-translated gloss)} & 88,424 & 96,450 & 184,874\\
\hline

$D_7$ & $D_4$ + Positive pairs {\small (original gloss - back-translated gloss)} & 144,009 & 96,450 & 240,459\\
\hline

$D_8$ & $D_5$ + Positive pairs {\small (original gloss - back-translated gloss)}  & 144,009 & 192,900 & 336,909\\
\hline

$D_9$ & $D_1$ + Positive pairs {\small (original context - back-translated gloss)} & 111,170 & 96,450 & 207,620\\
\hline

$D_{10}$ & $D_1$ + Pairs of cross relating the glosses against each other & 88,424 & 373,955 & 462,379\\
\hline

$D_{11}$ & $D_1$ (excluded pairs of functional words) & 54,878 & 92,730 & 147,608\\
\hline

$D_{12}$ & $D_1$ (only the pairs of the noun POS) & 36,487 & 37,998 & 74,485\\
\hline

$D_{13}$ & $D_1$ (only the pairs of the verb POS) & 18,178 & 54,945 & 73,123\\
\hline
\end{tabular}
\caption{\label{data-partitions}
The datasets that were used for the different experiments to fine-tune AraBERT on the TSV task.
}
\end{table*}

The original training dataset is not balanced with 55,585 positive pairs (36.6\%) and 96,450 negative pairs (63.4\%). To produce a more balanced dataset, we generated an additional 32,839 positive pairs by matching the original glosses with the new back-translated glosses increasing the number of positive pairs to 144,009. The 144,009 include 55,585 pairs from the original data, 55,585 pairs from back-translation and the added 32,839 pairs, resulting in a new dataset with 42.7\% positive and 57.3\% negative pairs. \\

\noindent \textbf{Observations on Google Translate:} First, although the quality of Google translations was generally acceptable, there are wrong translations. However, we did not make any improvements or revisions to these translations, as the goal of this paper is to measure whether automated back-translations can improve the accuracy of the trained models. Second, the output of the Google translation API was not always complete. In some cases it translates part of the input sentence. To overcome this challenge we used two techniques: 1) add special characters (.\#) at the end of each sentence, if the special characters were translated back, then we know the translation reached the end of the sentence, 2) compare the length of the original and back-translated sentences and if the difference is significant, then this is an indication of incomplete translations. Partial translations are reviewed manually.

\section{Experiments}
\label{sec:experiments}
This section presents 13 experiments to measure the impact of data augmentation using back-translation on TSV model accuracy. The first experiment uses the original ArabGlossBERT dataset,  $D_1$, (as a baseline), while the other experiments are conducted with different dataset configurations. In all experiments, we used the original test dataset 15,172 pairs (4,738 positive and 10,434 negative). Table~\ref{data-partitions} presents the training datasets that we used in the experiments.

\begin{table*}
\centering
\begin{tabular}
{|c|l||ll|l||ll|l||ll|l||lll|}

\hline
\multirow{4}{*}{\textbf{Dataset}} &
  \multicolumn{1}{c||}{\multirow{4}{*}{\textbf{Metric}}} &
  \multicolumn{2}{c|}{\multirow{2}{*}{\textbf{All POS}}} &
  \multirow{4}{*}{\textbf{\rotatebox[origin=c]{90}{Accuracy}}} &
  \multicolumn{2}{c|}{\multirow{2}{*}{\textbf{Noun}}} &
  \multirow{4}{*}{\textbf{\rotatebox[origin=c]{90}{Accuracy}}} &
  \multicolumn{2}{c|}{\multirow{2}{*}{\textbf{Verb}}} &
  \multirow{4}{*}{\textbf{\rotatebox[origin=c]{90}{Accuracy}}} &
  \multicolumn{2}{c|}{\multirow{2}{*}{\textbf{\makecell{Functional\\Words}}}} &
  \multirow{4}{*}{\textbf{\rotatebox[origin=c]{90}{Accuracy}}} \\  
 &
  \multicolumn{1}{c||}{} &
  \multicolumn{2}{c|}{} &
   &
  \multicolumn{2}{c|}{} &
   &
  \multicolumn{2}{c|}{} &
   &
  \multicolumn{2}{c|}{} &
   \\ \cline{3-4} \cline{6-7} \cline{9-10} \cline{12-13}
 &
  \multicolumn{1}{c||}{} &
  \multirow{2}{*}{\textbf{{\tiny \rotatebox[origin=c]{90}{Positive}}}} &
  \multirow{2}{*}{\textbf{{\tiny \rotatebox[origin=c]{90}{Negative}}}} &
   &
  \multirow{2}{*}{\textbf{{\tiny \rotatebox[origin=c]{90}{Positive}}}} &
  \multirow{2}{*}{\textbf{{\tiny \rotatebox[origin=c]{90}{Negative}}}} &
   &
  \multirow{2}{*}{\textbf{{\tiny \rotatebox[origin=c]{90}{Positive}}}} &
  \multirow{2}{*}{\textbf{{\tiny \rotatebox[origin=c]{90}{Negative}}}} &
   &
  \multirow{2}{*}{\textbf{{\tiny \rotatebox[origin=c]{90}{Positive}}}} &
  \multicolumn{1}{l|}{\multirow{2}{*}{\textbf{{\tiny \rotatebox[origin=c]{90}{Negative}}}}} &
   \\
 &
  \multicolumn{1}{c||}{} &
   &
   &
   &
   &
   &
   &
   &
   &
   &
   &
  \multicolumn{1}{l|}{} &
   \\ \hline
\makecell[c]{\vspace{-2mm} \textbf{D1} \\ \vspace{-0.5mm} {\tiny \textbf{Baseline}}  \vspace{-0.5mm} \\ {\scriptsize 152,035 pairs} } &
  \textbf{\begin{tabular}[c]{@{}l@{}}Precision\\ Recall\\ F1-Score\end{tabular}} &
  \begin{tabular}[c]{@{}l@{}}76\\ 66\\ 71\end{tabular} &
  \begin{tabular}[c]{@{}l@{}}85\\ 90\\ 88\end{tabular} &
  83 &
  \begin{tabular}[c]{@{}l@{}}75\\ 70\\ 72\end{tabular} &
  \begin{tabular}[c]{@{}l@{}}85\\ 88\\ 82\end{tabular} &
  82 &
  \begin{tabular}[c]{@{}l@{}}78\\ 65\\ 71\end{tabular} &
  \begin{tabular}[c]{@{}l@{}}85\\ 91\\ 88\end{tabular} &
  83 &
  \begin{tabular}[c]{@{}l@{}}63\\ 46\\ 53\end{tabular} &
  \multicolumn{1}{l|}{\begin{tabular}[c]{@{}l@{}}84\\ 92\\ 88\end{tabular}} &
  81 \\ \hline
\makecell[c]{\textbf{D2} \\ {\scriptsize 152,035 pairs}} &
  \textbf{\begin{tabular}[c]{@{}l@{}}Precision\\ Recall\\ F1-Score\end{tabular}} &
  \begin{tabular}[c]{@{}l@{}}81\\ 65\\ 72\end{tabular} &
  \begin{tabular}[c]{@{}l@{}}85\\ 93\\ 89\end{tabular} &
  84 &
  \begin{tabular}[c]{@{}l@{}}79\\ 68\\ 73\end{tabular} &
  \begin{tabular}[c]{@{}l@{}}85\\ 91\\ 88\end{tabular} &
  83 &
  \begin{tabular}[c]{@{}l@{}}82\\ 64\\ 72\end{tabular} &
  \begin{tabular}[c]{@{}l@{}}85\\ 94\\ 89\end{tabular} &
  84 &
  \begin{tabular}[c]{@{}l@{}}71\\ 36\\ 48\end{tabular} &
  \multicolumn{1}{l|}{\begin{tabular}[c]{@{}l@{}}82\\ 95\\ 88\end{tabular}} &
  81 \\ \hline
\makecell[c]{\textbf{D3} \\ {\scriptsize 152,035 pairs}} &
  \textbf{\begin{tabular}[c]{@{}l@{}}Precision\\ Recall\\ F1-Score\end{tabular}} &
  \begin{tabular}[c]{@{}l@{}}68\\ 52\\ 59\end{tabular} &
  \begin{tabular}[c]{@{}l@{}}80\\ 88\\ 84\end{tabular} &
  77 &
  \begin{tabular}[c]{@{}l@{}}65\\ 54\\ 59\end{tabular} &
  \begin{tabular}[c]{@{}l@{}}79\\ 85\\ 82\end{tabular} &
  75 &
  \begin{tabular}[c]{@{}l@{}}70\\ 52\\ 60\end{tabular} &
  \begin{tabular}[c]{@{}l@{}}80\\ 90\\ 85\end{tabular} &
  78 &
  \begin{tabular}[c]{@{}l@{}}55\\ 19\\ 29\end{tabular} &
  \multicolumn{1}{l|}{\begin{tabular}[c]{@{}l@{}}79\\ 95\\ 86\end{tabular}} &
  77 \\ \hline
\makecell[c]{\textbf{D4} \\ {\scriptsize 207,620 pairs}} &
  \textbf{\begin{tabular}[c]{@{}l@{}}Precision\\ Recall\\ F1-Score\end{tabular}} &
  \begin{tabular}[c]{@{}l@{}}80\\ 53\\ 64\end{tabular} &
  \begin{tabular}[c]{@{}l@{}}81\\ 94\\ 87\end{tabular} &
  81 &
  \begin{tabular}[c]{@{}l@{}}79\\ 55\\ 65\end{tabular} &
  \begin{tabular}[c]{@{}l@{}}80\\ 92\\ 86\end{tabular} &
  80 &
  \begin{tabular}[c]{@{}l@{}}81\\ 53\\ 64\end{tabular} &
  \begin{tabular}[c]{@{}l@{}}81\\ 94\\ 87\end{tabular} &
  81 &
  \begin{tabular}[c]{@{}l@{}}69\\ 23\\ 34\end{tabular} &
  \multicolumn{1}{l|}{\begin{tabular}[c]{@{}l@{}}80\\ 97\\ 88\end{tabular}} &
  79 \\ \hline
\makecell[c]{\textbf{D5} \\ {\scriptsize 304,070 pairs}} &
  \textbf{\begin{tabular}[c]{@{}l@{}}Precision\\ Recall\\ F1-Score\end{tabular}} &
  \begin{tabular}[c]{@{}l@{}}76\\ 57\\ 65\end{tabular} &
  \begin{tabular}[c]{@{}l@{}}82\\ 92\\ 87\end{tabular} &
  81 &
  \begin{tabular}[c]{@{}l@{}}77\\ 53\\ 63\end{tabular} &
  \begin{tabular}[c]{@{}l@{}}79\\ 92\\ 85\end{tabular} &
  80 &
  \begin{tabular}[c]{@{}l@{}}76\\ 62\\ 68\end{tabular} &
  \begin{tabular}[c]{@{}l@{}}84\\ 91\\ 87\end{tabular} &
  82 &
  \begin{tabular}[c]{@{}l@{}}70\\ 24\\ 36\end{tabular} &
  \multicolumn{1}{l|}{\begin{tabular}[c]{@{}l@{}}80\\ 97\\ 88\end{tabular}} &
  79 \\ \hline
\makecell[c]{\textbf{D6} \\ {\scriptsize 184,874 pairs}} &
  \textbf{\begin{tabular}[c]{@{}l@{}}Precision\\ Recall\\ F1-Score\end{tabular}} &
  \begin{tabular}[c]{@{}l@{}}76\\ 67\\ 71\end{tabular} &
  \begin{tabular}[c]{@{}l@{}}85\\ 90\\ 88\end{tabular} &
  \textbf{83} &
  \begin{tabular}[c]{@{}l@{}}76\\ 66\\ 71\end{tabular} &
  \begin{tabular}[c]{@{}l@{}}84\\ 89\\ \textbf{86}\end{tabular} &
  81 &
  \begin{tabular}[c]{@{}l@{}}76\\ 70\\ \textbf{73}\end{tabular} &
  \begin{tabular}[c]{@{}l@{}}87\\ 90\\ 88\end{tabular} &
  84 &
  \begin{tabular}[c]{@{}l@{}}71\\ 32\\ 44\end{tabular} &
  \multicolumn{1}{l|}{\begin{tabular}[c]{@{}l@{}}82\\ 96\\ 88\end{tabular}} &
  81 \\ \hline
\makecell[c]{\textbf{D7} \\ {\scriptsize 240,459 pairs}} &
  \textbf{\begin{tabular}[c]{@{}l@{}}Precision\\ Recall\\ F1-Score\end{tabular}} &
  \begin{tabular}[c]{@{}l@{}}79\\ 56\\ 66\end{tabular} &
  \begin{tabular}[c]{@{}l@{}}82\\ 93\\ 87\end{tabular} &
  81 &
  \begin{tabular}[c]{@{}l@{}}77\\ 57\\ 66\end{tabular} &
  \begin{tabular}[c]{@{}l@{}}81\\ 91\\ 86\end{tabular} &
  80 &
  \begin{tabular}[c]{@{}l@{}}80\\ 58\\ 67\end{tabular} &
  \begin{tabular}[c]{@{}l@{}}83\\ 93\\ 88\end{tabular} &
  82 &
  \begin{tabular}[c]{@{}l@{}}71\\ 17\\ 17\end{tabular} &
  \multicolumn{1}{l|}{\begin{tabular}[c]{@{}l@{}}79\\ 98\\ 98\end{tabular}} &
  79 \\ \hline
\makecell[c]{\textbf{D8} \\ {\scriptsize 336,909 pairs}} &
  \textbf{\begin{tabular}[c]{@{}l@{}}Precision\\ Recall\\ F1-Score\end{tabular}} &
  \begin{tabular}[c]{@{}l@{}}80\\ 54\\ 65\end{tabular} &
  \begin{tabular}[c]{@{}l@{}}81\\ 94\\ 87\end{tabular} &
  81 &
  \begin{tabular}[c]{@{}l@{}}79\\ 55\\ 65\end{tabular} &
  \begin{tabular}[c]{@{}l@{}}80\\ 92\\ 86\end{tabular} &
  80 &
  \begin{tabular}[c]{@{}l@{}}81\\ 53\\ 64\end{tabular} &
  \begin{tabular}[c]{@{}l@{}}81\\ 94\\ 87\end{tabular} &
  81 &
  \begin{tabular}[c]{@{}l@{}}69\\ 23\\ 34\end{tabular} &
  \multicolumn{1}{l|}{\begin{tabular}[c]{@{}l@{}}80\\ 97\\ 88\end{tabular}} &
  79 \\ \hline

\makecell[c]{\textbf{D9} \\ {\scriptsize 207,620 pairs}} &
  \textbf{\begin{tabular}[c]{@{}l@{}}Precision\\ Recall\\ F1-Score\end{tabular}} &
  \begin{tabular}[c]{@{}l@{}}78\\ 63\\ 70\end{tabular} &
  \begin{tabular}[c]{@{}l@{}}84\\ 92\\ 88\end{tabular} &
  \textbf{83} &
  \begin{tabular}[c]{@{}l@{}}77\\ 62\\ 69\end{tabular} &
  \begin{tabular}[c]{@{}l@{}}83\\ 91\\ \textbf{86}\end{tabular} &
  81 &
  \begin{tabular}[c]{@{}l@{}}78\\ 66\\ \textbf{72}\end{tabular} &
  \begin{tabular}[c]{@{}l@{}}86\\ 92\\ 88\end{tabular} &
  84 &
  \begin{tabular}[c]{@{}l@{}}73\\ 31\\ 43\end{tabular} &
  \multicolumn{1}{l|}{\begin{tabular}[c]{@{}l@{}}81\\ 96\\ 88\end{tabular}} &
  81 \\ \hline
\makecell[c]{\textbf{D10} \\ {\scriptsize 462,379 pairs}} &
  \textbf{\begin{tabular}[c]{@{}l@{}}Precision\\ Recall\\ F1-Score\end{tabular}} &
  \begin{tabular}[c]{@{}l@{}}71\\ 51\\ 59\end{tabular} &
  \begin{tabular}[c]{@{}l@{}}80\\ 90\\ 85\end{tabular} &
  78 &
  \begin{tabular}[c]{@{}l@{}}70\\ 50\\ 58\end{tabular} &
  \begin{tabular}[c]{@{}l@{}}78\\ 89\\ 83\end{tabular} &
  76 &
  \begin{tabular}[c]{@{}l@{}}71\\ 54\\ 61\end{tabular} &
  \begin{tabular}[c]{@{}l@{}}81\\ 90\\ 85\end{tabular} &
  79 &
  \begin{tabular}[c]{@{}l@{}}66\\ 19\\ 30\end{tabular} &
  \multicolumn{1}{l|}{\begin{tabular}[c]{@{}l@{}}79\\ 97\\ 87\end{tabular}} &
  78 \\ \hline

  \makecell[c]{\textbf{D11} \\ {\scriptsize 147,750 pairs}} &
  \textbf{\begin{tabular}[c]{@{}l@{}}Precision\\ Recall\\ F1-Score\end{tabular}} &
  \begin{tabular}[c]{@{}l@{}}80\\ 54\\ 65\end{tabular} &
  \begin{tabular}[c]{@{}l@{}}81\\ 94\\ 87\end{tabular} &
  81 &
  \begin{tabular}[c]{@{}l@{}}79\\ 55\\ 65\end{tabular} &
  \begin{tabular}[c]{@{}l@{}}80\\ 92\\ 86\end{tabular} &
  80 &
  \begin{tabular}[c]{@{}l@{}}81\\ 53\\ 64\end{tabular} &
  \begin{tabular}[c]{@{}l@{}}81\\ 94\\ 87\end{tabular} &
  81 &
 \begin{tabular}[c]{@{}l@{}}\\ \\ \end{tabular} &
  \multicolumn{1}{l|}{\begin{tabular}[c]{@{}l@{}}\\ \\ \end{tabular}} &
   \\ \hline
   
\makecell[c]{\textbf{D12} \\ {\scriptsize 74,485 pairs}} &
  \textbf{\begin{tabular}[c]{@{}l@{}}Precision\\ Recall\\ F1-Score\end{tabular}} &
  \begin{tabular}[c]{@{}l@{}}\\ \\ \end{tabular} &
  \begin{tabular}[c]{@{}l@{}}\\ \\ \end{tabular} &
   &
  \begin{tabular}[c]{@{}l@{}}80\\ 60\\ 69\end{tabular} &
  \begin{tabular}[c]{@{}l@{}}82\\ 92\\ 87\end{tabular} &
  81 &
  \begin{tabular}[c]{@{}l@{}}\\ \\ \end{tabular} &
  \begin{tabular}[c]{@{}l@{}}\\ \\ \end{tabular} &
   &
  \begin{tabular}[c]{@{}l@{}}\\ \\ \end{tabular} &
  \multicolumn{1}{l|}{\begin{tabular}[c]{@{}l@{}}\\ \\ \end{tabular}} &
   \\ \hline
\makecell[c]{\textbf{D13} \\ {\scriptsize 73,123 pairs}} &
  \textbf{\begin{tabular}[c]{@{}l@{}}Precision\\ Recall\\ F1-Score\end{tabular}} &
  \begin{tabular}[c]{@{}l@{}}\\ \\ \end{tabular} &
  \begin{tabular}[c]{@{}l@{}}\\ \\ \end{tabular} &
   &
  \begin{tabular}[c]{@{}l@{}}\\ \\ \end{tabular} &
  \begin{tabular}[c]{@{}l@{}}\\ \\ \end{tabular} &
   &
  \begin{tabular}[c]{@{}l@{}}74\\ 62\\ 68\end{tabular} &
  \begin{tabular}[c]{@{}l@{}}84\\ 90\\ 87\end{tabular} &
   81&
  \begin{tabular}[c]{@{}l@{}}\\ \\ \end{tabular} &
  \multicolumn{1}{l|}{\begin{tabular}[c]{@{}l@{}}\\ \\ \end{tabular}} &
   \\ \hline
\end{tabular}
\caption{\label{experiments-result}Results, expressed as percentage, of the experiments for fine-tuning AraBERT on different combinations of the original ArabGlossBERT and augmented datasets.}
\end{table*}

In all experiments we fine-tuned AraBERTv2 (aubmindlab/bert-base-arabertv02, CC-BY-SA) using the following hyperparameters: $\eta=2e^{-5}$, batch size $B=16$, max sequence length of 512, warm-up steps 1,412 and number of epochs 4.

The results of the 13 experiments are presented in Table \ref{experiments-result}, which includes precision, recall, F1-score, and accuracy. The results are presented at the POS tag level and overall. Also, note that the test dataset is the same test set used in the original ArabGlossBERT dataset because we consider ArabGlossBERT as a baseline. In the next sub-sections, we elaborate on each experiment.

\subsection{\texorpdfstring{Experiment 1: $D_1$ Dataset (Baseline)}{}}
This experiment is the baseline for results comparison. We used the original dataset ArabGlossBERT, $D_1$, without any augmentation and achieved the same results (83\% accuracy) as reported in (\citealp{al2022arabglossbert}). Additionally, we evaluated the model performance per POS tag since the tokens are annotated with the POS tags (noun, verb, and functional words). While the accuracy across all tags is very similar (Table \ref{experiments-result},), we observe a big difference in the Positive pair F1-score. For the functional words, the F1-score for Positive pairs is only 53\%, compared to 72\% and 71\% for the nouns and verbs, respectively. We will notice this trend across all experiments, since functional words are highly polysemous (e.g., the preposition (\Ar{في} / in) has ten different glosses), and their glosses represent function and use in the sentence, rather than semantics.

\subsection{\texorpdfstring{Experiment 2: $D_2$ Dataset}{}}
The idea of this experiment is to use a learning signal by marking up the target word, in its context, with an open-close tag (<token>Target</token>) to emphasize the model learning of the target word. Thus, the dataset $D_2$ is the same as $D_1$ but with a learning signal surrounding the target words. This experiment is the same experiment conducted in \cite{al2022arabglossbert} and we achieved the same results (84\% accuracy). Overall, we see a 1\% increase by using $D_2$ over $D_1$. We note that $D_2$ is the only dataset with the target signal added.

\subsection{\texorpdfstring{Experiment 3: $D_3$ Dataset}{}}
This experiment evaluates the model performance using $D_3$, which contains the back-translated context and the original gloss pairs (152,035). As shown in Table~\ref{experiments-result}, the overall accuracy creased from 83\% on $D_1$ to 77\%  on $D_3$. The 6\% drop in the accuracy illustrates that the quality of the back-translations is acceptable as an augmentation to the original data.

\subsection{\texorpdfstring{Experiment 4: $D_4$ Dataset}{}}
$D_4$ is original dataset $D_1$ in addition to the 55,585 Positive back-translated pairs. The motivation of adding the Positive back-translated pairs is to balance the original dataset, $D_1$. Recall that $D_1$ contains 55,585 Positive pairs (36.6\%) and 96,450 Negative pairs (63.4\%) and by adding the Positive back-translated pairs, $D_4$ size increases to 207,620 pairs, among which 111,170 (53.5\%) are positive pairs. Table~\ref{experiments-result} shows that this data configuration did not improve the model performance. On the contrary, the accuracy dropped by 2\% compared to $D_1$ (baseline). We also note that the F1-score dropped from 71\% to 64\% for Positive pairs, and from 88\% to 87\% for Negative pairs.

\subsection{\texorpdfstring{Experiment 5: $D_5$ Dataset}{}}
$D_5$ consists of the original dataset $D_1$ in addition to its back-translation dataset $D_3$. Although $D_5$ is large (304,070 pairs), its accuracy is 81\%, which is 2\% lower than the baseline. 

\subsection{\texorpdfstring{Experiment 6: $D_6$ Dataset}{}}
The $D_6$ dataset used in this experiment contains the original dataset $D_1$, in addition to 32,839 Positive pairs that we generated by paring an original gloss with its back-translation. We achieved the same accuracy as the baseline (83\%), but we believe that the fine-tuned model on $D_6$ is a little better than the baseline model for two reasons. First, the the F1-score for \textit{noun} Negative pairs increased by 4\% compared to the baseline to 86\%, and the F1-score for \textit{verb} Positive pairs increased by 2\% to 73\%. Second, since the training dataset is larger it is assumed to cover more vocabulary.   

\subsection{\texorpdfstring{Experiments 7-8: $D_7$ and $D_8$ Datasets}{}}
Although we increased the size of datasets in these two experiments, their model accuracy and F1-scores are very similar, but lower compared with the baseline. $D_7$ contains the original dataset, the Positive back-translated pairs and the Positive glosses with their back-translations. With this data, we increased the Positive pairs to be 144,009 (60\%) of the dataset. In experiment 8 we used $D_8$, which contains the original dataset, all back-translation pairs, and the Positive gloss-gloss pairs. 

\subsection{\texorpdfstring{Experiment 9: $D_9$ Dataset}{}}
$D_9$ contains $D_1$ and the 55,585 Positive pairs that we produced by pairing the original context with their back-translated gloss. The Positive pairs in $D_9$ account for 53.5\% of the dataset. This data configuration achieved the same as the baseline (83\% accuracy). Although the performance is same as the baseline, we see similar behaviour and we conclude the same as we did on the dataset $D_6$.  

\subsection{\texorpdfstring{Experiment 10: $D_{10}$ Dataset}{}}
In this experiment we did not use back-translation. However, we augmented the original dataset $D_1$ such that, the set of glosses of a certain lemma are cross-related and the resulting pairs are considered Negative pairs. In this way, we were able to generate 32,839 Positive pairs and 277,505 Negative pairs, a total of 310,344 pairs. We augmented these pairs with $D_1$ resulting in 462,379 pairs. Notice that this is the hardest dataset to model because some negative pairs are generated at the lemma level and are harder to distinguish from their positive counterparts. The idea is to fine-tune a model to be more sensitive in distinguishing positive and negative pairs, which as expected resulted in the lowest  performance (78\% accuracy) compared to other models. 

\subsection{\texorpdfstring{Experiment 11: $D_{11}$ Dataset}{}}
The goal of this experiment is to fine-tune a model excluding all pairs that are labeled with functional words. Functional words such as (\Ar{إذا ،من، على، في، إلى}) play the role of particles rather than providing core semantics. Additionally, they are frequently used in contexts and are highly polysemous. We fine-tuned a model with the $D_{11}$ dataset, which is the same as the original dataset $D_1$, but it excludes 4,427 pairs of functional words. However, the performance did not improve compared to the baseline. This illustrates that keeping the pairs of functional words is better than excluding them. 

\subsection{\texorpdfstring{Experiment 12-13: $D_{12}$ and $D_{13}$ Datasets}{}}
The goal of these two experiments is to evaluate the pairs labeled with \textit{nouns} and \textit{verbs} separately. $D_{12}$ contains 74,485 pairs, in which targets are \textit{nouns} only, and $D_{13}$ contains 73,123 pairs with \textit{verb} targets. We fine-tuned two separate models for each of the datasets and achieved similar accuracy and F1-scores, however, the performance is slightly lower compared to the baseline. Nevertheless, since both $D_{12}$ and $D_{13}$ achieved similar results, we believe that fine-tuning the model on data with both POS tags allows for cross learning and in turn yields better performance.

\section{Discussion and Conclusions}
\label{sec:discussionconclusion}
We presented an approach to improve Arabic TSV using automatic back-translation.
We augmented an existing Arabic TSV dataset, ArabGlossBERT, by doubling its size with back-translated data using Google Translate API. To measure the impact of the data augmentation, we presented 13 experiments with different data configurations. Although we did not outperform the overall performance of the baseline model, we did observe that some experiments such as $D_6$ outperformed the baseline on \textit{noun} positive pair and \textit{verb} negative pair classification. Overall, our results are close to the results presented in \cite{lin2021context}, which used back-translation augmentation for English TSV and achieved only 2\% F1-score improvement. Nevertheless, we would like to note the following findings:

\begin{itemize}

\item Fine-tuning a BERT model using only the back-translation pairs achieved 77\% accuracy (experiment 3), which is only 6\% less than the baseline accuracy. This illustrates that the quality of automatic translations of glosses and contexts is not high but is generally acceptable.

\item The different augmentations to the original dataset achieved between 78\% to 83\% accuracy (see experiments 4-9), but it did not outperform the baseline model. At the same time, augmentation did not harm the performance since the results are comparable to the baseline. 
Nevertheless, experiments 6 and 9 have illustrated a small improvement in the F1-scores for \textit{noun} and \textit{verb} POS. In addition, because $D_6$ and $D_9$ are larger than the baseline $D_1$ dataset, the fine-tuned models are assumed to cover a larger vocabulary and more contexts.

\item Looking at the F1-scores, we note that the Positive pairs are always lower than the Negative pairs in all experiments and for all POS categories. This means that all models are less accurate at predicting Positive pairs. Although we tried to augment the dataset by increasing the number of Positive pairs, the F1-scores did not improve. 

\item In our attempts to fine-tune different models for each POS category, we found that: (1) excluding the pairs of functional words from the dataset (experiment 11) did not improve the performance, and (2) fine-tuning a model for all POS categories allows for cross learning from different POS tagged targets and yields better performance than fine-tuning separate models for \textit{nouns} and \textit{verbs} (experiments 12-13).  

\end{itemize}

\section{Limitations and Future Works}
\label{sec:future_work}
Our data augmentation as well as the experiments are based on (1) the quality of Google Translate API, (2) the quality of the glosses and contexts in the ArabGlossBERT training dataset, and most importantly on (3) the quality and coverage of the ArabGlossBERT test dataset. Although the quality of machine translation is limited, the goal of this paper is to measure whether such limited translations can improve the accuracy of the TSV fine-tuned models. Additionally, the quality of the glosses and contexts in the ArabGlossBERT training dataset cannot be improved since they originated from Arabic lexicons. However, we believe that enriching the ArabGlossBERT by collecting more pairs from Arabic lexicons (i.e., building a rich Arabic sense inventory) will empower research on TSV and WSD tasks. More importantly, all experiments conducted in this paper used the ArabGlossBERT test dataset. Since there are no other testing datasets or benchmarks, the evaluation of our fine-tuned models is limited  based on the quality and coverage of the ArabGlossBERT test dataset.

Next, we plan to develop another test dataset to evaluate our models and their generalizability. We plan to further explore other approaches for WSD task such as ranking of glosses, rather than addressing the WSD task through TSV.

\section*{Acknowledgment}
\label{sec:ack}
We would like to thank Taymaa Hammouda and Ala Omar for the technical support on many aspects of this research.

\bibliography{paper}

\end{document}